\documentclass[runningheads]{llncs}
\usepackage{graphicx}
\usepackage{amsmath,amssymb} 
\usepackage{color}
\usepackage[width=122mm,left=12mm,paperwidth=146mm,height=193mm,top=12mm,paperheight=217mm]{geometry}

\usepackage{booktabs}
\usepackage{subfigure}
\usepackage{array}

\begin{document}
\pagestyle{headings}
\mainmatter

\title{Recurrent Fusion Network for Image Captioning} 

\titlerunning{Recurrent Fusion Network for Image Captioning}

\authorrunning{Wenhao Jiang, Lin Ma, Yu-Gang Jiang, Wei Liu, Tong Zhang}

\author{Wenhao Jiang$^{1}$, Lin Ma$^{1}$, Yu-Gang Jiang$^{2}$, Wei Liu$^{1}$, Tong Zhang$^{1}$} 
\institute{$^{1}$Tencen AI Lab,
$^{2}$Fudan University\\
\email{ \{cswhjiang, forest.linma\}@gmail.com, ygj@fudan.edu.cn, wl2223@columbia.edu, tongzhang@tongzhang-ml.org}
}



\maketitle

\begin{abstract}
Recently, much advance has been made in image captioning, and an encoder-decoder framework has been adopted by all the state-of-the-art models. Under this framework, an input image is encoded by a convolutional neural network (CNN) and then translated into natural language with a recurrent neural network (RNN). The existing models counting on this framework merely employ one kind of CNNs, \textit{e.g.}, ResNet or Inception-X, which describe image contents from only one specific view point. Thus, the semantic meaning of an input image cannot be comprehensively understood, which restricts  the performance of captioning. In this paper, in order to exploit the complementary information from multiple encoders, we propose a novel Recurrent Fusion Network (RFNet) for tackling image captioning. The fusion process in our model can exploit the interactions among the outputs of the image encoders and then generate new compact yet informative representations for the decoder. Experiments on the MSCOCO dataset demonstrate the effectiveness of our proposed RFNet, which sets a new state-of-the-art for image captioning.
\end{abstract}

\section{Introduction}
Captioning \cite{venugopalan2015sequence,krishna2017dense,chen2018less,wang2018reconstruction,wang2018bidirectional,karpathy2015deep,vinyals2016show}, a task to describe images/videos with natural sentences automatically, has been an active research topic in computer vision and machine learning. Generating natural descriptions of images is very useful in practice. For example, it can improve the quality of image retrieval by discovering salient contents and help visually impaired people understand image contents. 

Even though a great success has been achieved in object recognition \cite{he_eccv2014,krizhevsky2012imagenet}, describing images with natural sentences is still a very challenging task. Image captioning models need to have a thorough understanding of an input image and capture the complicated relationships among objects. Moreover, they also need to capture the interactions between images and languages and thereby translate image representations into natural sentences.

The encoder-decoder framework, with its advance in machine translation \cite{cho2014learning,bahdanau2014neural}, has demonstrated promising performance in image captioning \cite{vinyals2016show,xu2015show,chen2018regularizing}. This framework consists of two parts, namely an encoder and a decoder. The encoder is usually a convolutional neural network (CNN), while the decoder is a recurrent neural network (RNN). The encoder is used to extract image representations, based on which the decoder is used to generate the corresponding captions. Usually, a pre-trained CNN targeting for image classification is leveraged to extract image representations.

All existing models employ only one encoder, so the performance heavily depends on the expressive ability of the deployed CNN. Fortunately, there are quite a few well-established CNNs, \textit{e.g.}, ResNet~\cite{he2016deep}, Inception-X~\cite{szegedy2016rethinking,szegedy2017inception}, \textit{etc}. It is natural to improve  captioning performance by extracting diverse representations with multiple encoders, which play a complementary role in fully depicting and characterizing semantic information of images. However, to the best of our knowledge, there is no model that considers to exploit the complementary behaviors of multiple encoders for image captioning.

\begin{figure}[t] 
\begin{center}
\includegraphics[width=1\linewidth]{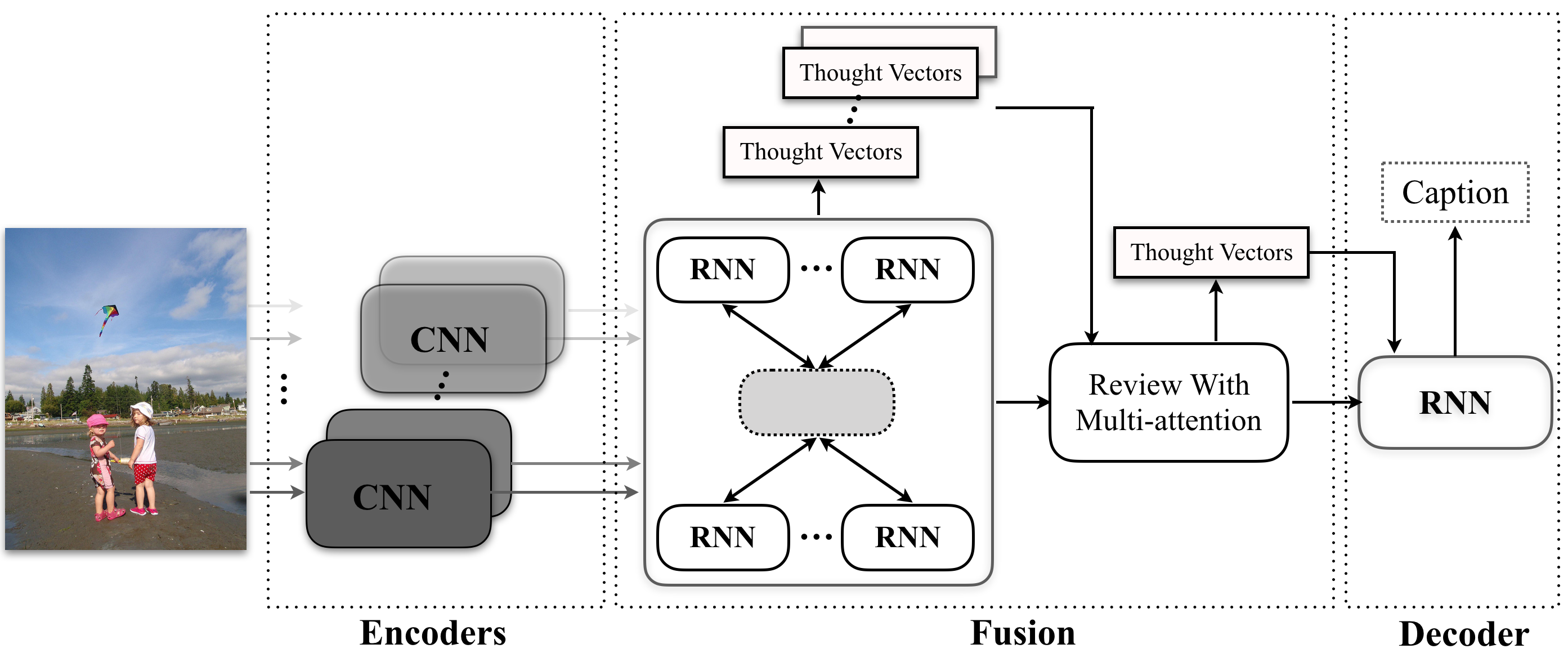}
\end{center}
   \caption{The framework of our proposed RFNet. Multiple CNNs are employed as the encoders and a recurrent fusion procedure is inserted after the encoders to form better representations for the decoder. The fusion procedure consists of two stages. The first stage exploits interactions among the representations from multiple CNNs to generate multiple sets of thought vectors. The second stage performs multi-attention on the sets of thought vectors from the first stage, and generates a new set of thought vectors for the decoder.}
\label{fig:framework}
\end{figure} 

In this paper, to exploit complementary information from multiple encoders, we propose a Recurrent Fusion Network (RFNet) for image captioning. Our framework, as illustrated in Fig.~\ref{fig:framework}, introduces a fusion procedure between the encoders and decoder. Multiple CNNs, served as the encoders, can provide diverse and more comprehensive descriptions of the input image. The fusion procedure performs a given number of RNN steps and outputs the hidden states as thought vectors. Our fusion procedure consists of two stages. The first stage contains multiple components, and each component processes the information from one encoder. The interactions among the components are captured to generate thought vectors. Hence, each component can communicate with the other components and extract complementary information from them. The second stage compresses the outputs  of the first stage into one set of thought vectors. During this procedure, the interactions among the sets of thought vectors are further exploited. Thus, useful information is absorbed into the final thought vectors, which will be used as input of the attention model in the decoder.
The intuition behind our proposed RFNet is to fuse all the information encoded by  multiple encoders and produce thought vectors that are more comprehensive and representative than the original ones. 

\section{Related Works}
\subsection{Encoder-Decoder Methods for Image Captioning}
Recently, inspired by advance in machine translation, the encoder-decoder framework \cite{sutskever2014sequence,cho2014learning}  has also been introduced to image captioning \cite{vinyals2015show}. In this framework, a CNN pre-trained on an image classification task is used as the encoder, while an RNN is used as the decoder to translate the information from the encoder into natural sentences. This framework is simple and elegant, and several extensions have been proposed. In \cite{xu2015show}, a spatial attention mechanism was introduced. The model using this attention mechanism could determine which subregions should be focused on automatically at each time step. In~\cite{chen2017sca}, channel- wise attention was proposed to modulates the sentence generation context in multi-layer feature maps. In \cite{yang2016review}, ReviewNet was proposed. The review steps can learn the annotation vectors and initial states for the decoder, which are more representative than those generated by the encoder directly. In~\cite{jiang2018learning}, a guiding network which models attribute properties of input images was introduced for the decoder. In \cite{lu2016knowing}, the authors observed that the decoder does not need visual attendance when predicting non-visual words, and hence proposed an adaptive attention model that attends to the image or to the visual sentinel automatically at each time step.

Besides, several approaches that introduce useful information into this framework have been proposed to improve the image captioning performance. In \cite{wu2016value}, the word occurrence prediction was treated as a multi-label classification problem. And a region-based multi-label classification framework was proposed to extract visually semantic information. This prediction is then used to initialize a memory cell of a long short-term memory (LSTM) model. Yao $et~al.$ improved this procedure and discussed different approaches to incorporate word occurrence predictions into the decoder~\cite{yao2016boosting}.

Recently, in order to optimize the non-differentiable evaluation metrics directly, policy gradient methods for reinforcement learning are employed to train the encoder-decoder framework. In~\cite{rennie2016self}, the cross-entropy loss was replaced by negative CIDEr score~\cite{vedantam2015cider}. This system was then trained with the REINFORCE algorithm~\cite{williams1992simple}, which significantly improved the performance. Such a training strategy can be leveraged to improve the performance of all existing models under the encoder-decoder framework.

\subsection{Encoder-Decoder Framework with Multiple Encoders or Decoders}
In \cite{Luong2016}, multi-task learning (MTL) was combined with sequence-to-sequence learning with multiple encoders or decoders. In the multi-task sequence-to-sequence learning, the encoders or decoders are shared among different tasks. The goal of \cite{Luong2016} is to transfer knowledge among tasks to improve the performance. For example, the tasks of translation and image captioning can be formulated together as a model with only one decoder. The decoder is shared between both tasks and responsible for translating from both image and source language. Both tasks can benefit from each other. A similar structure was also exploited in \cite{firat2016multi} to perform multi-lingual translation. In this paper, we propose a model to combine representations from multiple encoders for the decoder. In \cite{Luong2016,firat2016multi}, the inputs of the encoders are different. But in our model, they are the same. Our goal is to leverage complementary information from different encoders to form better representations for the decoder.

\subsection{Ensemble and Fusion}
Our proposed RFNet also relates to information fusion, multi-view learning \cite{xu2013survey}, and ensemble learning \cite{zhou2012ensemble}. Each representation extracted from an individual image CNN can be regarded as an individual view depicting the input image.  Combining different representations with diversity is a well-known technique to improve the performance. The combining process can occur at the input, intermediate, and output stage of the target model. For the input fusion, the simplest way is to concatenate all the representations and use the concatenation as the input of the target model. This method usually leads to limited improvements. For the output fusion, the results of base learners for individual views are combined to form the final results. The common ensemble technique in image captioning is regarded as an output fusion technique, combining the output of the decoder at each time step~\cite{vinyals2015show,yang2016review,rennie2016self}.  For the intermediate fusion, the representations from different views are preprocessed by exploiting the relationships among them to form the input for the target model. Our proposed RFNet can be regarded as a kind of intermediate fusion methods.

\section{Background}
To provide a clear description of our method,  we present a short review of the encoder-decoder framework for image captioning in this section. 

\subsection{Encoder}
Under the encoder-decoder framework for image captioning, a CNN pre-trained for an image classification task is usually employed as the encoder to extract the global representation and subregion representations of an input image. The global representation is usually the output of a fully-connected layer, while subregion representations are usually the outputs of a convolutional layer. The extracted global representation and subregion representations are denoted as $\mathbf{a}_0$ and $A = \{\mathbf{a}_1, \dots, \mathbf{a}_k\}$, respectively, where $k$ denotes the  number of subregions.

\subsection{Decoder}
 Given the image representations $\mathbf{a}_0$ and  $A$, a decoder, which is usually a gated recurrent unit (GRU) \cite{chung2014empirical} or long short-term memory (LSTM)~\cite{hochreiter1997long}, is employed to translate an input image into a natural sentence. In this paper, we use LSTM equipped with an attention mechanism as the basic unit of the decoder.

Recall that an LSTM with an attention mechanism is a function that outputs the results based on the current hidden state, current input, and context vector. The context vector is the weighted sum of elements from $A$, with the weights determined by an attention model. We adopt the same LSTM used in \cite{xu2015show} and express the LSTM unit with the attention strategy  as follows:
\begin{align}
\left( \begin{array}{c}
\mathbf{i}_{t}  \\
\mathbf{f}_{t} \\
\mathbf{o}_{t} \\
\mathbf{g}_{t} \\
\end{array} \right) &=
\left( \begin{array}{c}
\sigma  \\
\sigma \\
\sigma \\
\textrm{tanh} \\
\end{array} \right)
\mathbf{T}
\left( \begin{array}{c}
\mathbf{H}_t  \\
\mathbf{z}_{t-1} \\
\end{array} \right), \\
\mathbf{c}_t &= \mathbf{f}_t \odot \mathbf{c}_{t-1} + \mathbf{i}_{t}\odot \mathbf{g}_t, \\
\mathbf{h}_t &= \mathbf{o}_t \odot \textrm{tanh}(\mathbf{c}_t),
\end{align}
where $\mathbf{i}_t$, $\mathbf{f}_t$, $\mathbf{c}_t$, $\mathbf{o}_t$, and $\mathbf{h}_t$ are the input gate, forget gate, memory cell, output gate, and hidden state of the LSTM, respectively. Here, 
\begin{align}
\mathbf{H}_t=	\left[ \begin{array}{c}
\mathbf{x}_t  \\
\mathbf{h}_{t-1}  \\
\end{array} \right]
\end{align}
is the concatenation of input $\mathbf{x}_t$ at time step $t$ and hidden state $\mathbf{h}_{t-1}$, $\mathbf{T}$ is a linear transformation operator. $\mathbf{z}_t $ is the context vector, which is the output of the attention model $f_{\textrm{att}}({A}, \mathbf{h}_{t-1})$. Specifically, 
\begin{align}
	e_{ti} = \text{sim}(\mathbf{a}_i, \mathbf{h}_{t-1}), \
	\alpha_{ti} = \frac{\exp(e_{ti})}{\sum_{j=1}^{k} \exp(e_{tj})}, \ \textrm{and} \
	\mathbf{z}_t = \sum_{i=1}^{k}  \alpha_{ti}  \mathbf{a}_i, \nonumber
\end{align}
where $\text{sim}(\mathbf{a}_i, \mathbf{h}_t) $ is a function to measure the similarity between $\mathbf{a}_i$ and $\mathbf{h}_t$, which is usually realized by a multilayer perceptron (MLP). In this paper, we use the shorthand notation
\begin{align}
	[\mathbf{h}_t,\mathbf{c}_t] = \textrm{LSTM}(\mathbf{H}_t, f_{\textrm{att}}(A, \mathbf{h}_{t-1}))
\end{align}
for the above equations.

The purpose of image captioning is to generate a caption $\mathcal{C} = (y_1,y_2,\cdots, y_N)$  for one given image $\mathcal{I}$. The objective adopted is usually to minimize a negative log-likelihood:
\begin{align}\label{obj_lm}
\mathcal{L} = -\log\ p(\mathcal{C}|\mathcal{I}) = -\sum_{t=0}^{N-1}  \log p(y_{t+1}|y_{t}),	
\end{align}
where
$p(y_{t+1}|y_{t}) = \textrm{Softmax}(\mathbf {W}\mathbf{h}_t)$  and $\mathbf{h}_t$ is computed by setting $\mathbf{x}_t = \mathbf{E}\mathbf{y}_{t}$.
Here, $\mathbf{W}$ is a matrix for linear transformation and $y_0$ is the sign for the start of sentences.  $\mathbf{E}\mathbf{y}_{t}$ denotes the distributed representation of the word $\mathbf{y}_{t}$, in which $\mathbf{y}_{t}$ is the one-hot representation for word $y_{t}$ and $\mathbf{E}$ is the word embedding matrix.

\section{Our Method}  

In this section, we propose our Recurrent Fusion Network (RFNet) for image captioning.
The fusion process in RFNet consists of two stages. The first stage combines the representations from multiple encoders to form multiple sets of thought vectors, which will be compressed into one set of thought vectors in the second stage. The goal of our model is to generate more representative thought vectors for the decoder. Two special designs are adopted: 1) employing interactions among components in the first stage; 2) reviewing the thought vectors from the previous stage in the second stage. We will describe the  details of our RFNet in this section and analyze our design in the experiments section.
 
\subsection{The Architecture}
\begin{figure}[t]
\begin{center}
\includegraphics[width=1\linewidth]{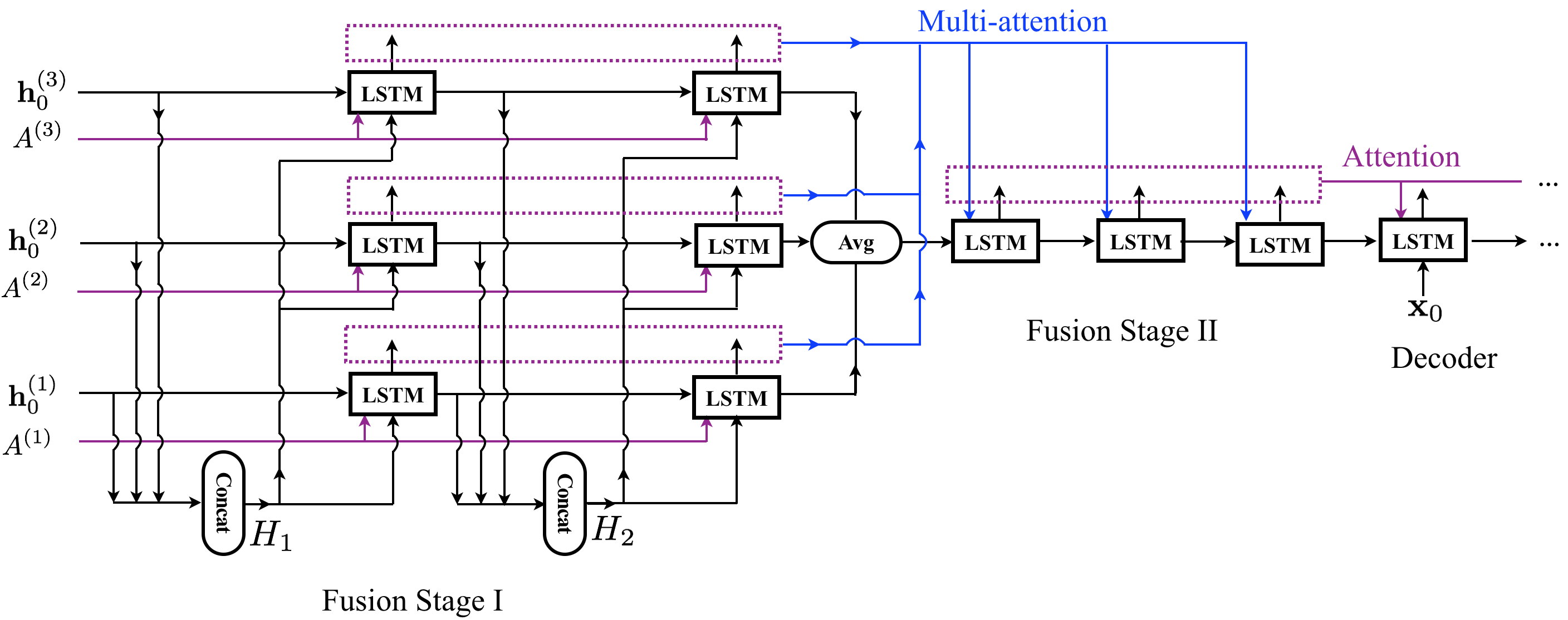}
\end{center}
   \caption{An example with $M=3$, $T_1 = 2$, and $T_2 = 3$ for illustrating our proposed RFNet (encoders are omitted). The fusion stage I contains $M$ review components. The input of each review component at each time step is the concatenation of hidden states from all components at the previous time step. Each review component outputs the hidden state as a thought vector. The fusion stage II is a review component that performs the multi-attention mechanism on the multiple sets of thought vectors from the fusion stage I. The parameters of LSTM units in the fusion procedure are all different. The purple rectangles indicate the sets of thought vectors. Moreover, in order to make it clear, the discriminative supervision is not presented.}
\label{fig:funsion_step}
\end{figure}


In our model, $M$ CNNs serve as the encoders. The global representation and subregion representations extracted from the $m$-th CNN are denoted as $\mathbf{a}_0^{(m)}$ and $A^{(m)} = \{\mathbf{a}_1^{(m)}, \dots, \mathbf{a}_{k_m}^{(m)}\}$, respectively. 

The framework of our proposed RFNet is illustrated in Fig.~\ref{fig:funsion_step}. The fusion procedure of RFNet consists of two stages, specifically fusion stages I and II.  Both stages perform a number of RNN steps with attention mechanisms and output hidden states as the thought vectors. The numbers of steps in stages I and II are denoted as $T_1$ and $T_2$, respectively. The hidden states from stages I and II are regarded as the thought vectors. The thought vectors of stage I will be used as the input of the attention model of stage II. The hidden states and memory cells after the last step of fusion stage I are aggregated to form the initial hidden state and the memory cell for fusion stage II. The thought vectors generated by stage II will be used in the decoder. RFNet is designed to capture the interactions among the components in stage I, and compress  the $M$ sets of thought vectors to generate more compact and informative ones in stage II. The details will be described in the following subsections.

\subsection{Fusion Stage I}

Fusion stage I takes $M$ sets of annotation vectors as inputs and generates $M$ sets of thought vectors, which will be aggregated into one set of thought vectors in fusion stage II. This stage contains $M$ review components. In order to capture the interactions among the review components,  each review component needs to know what has been generated by all the components at the previous time step. 
 
We denote the hidden state and memory cell of the $m$-th review component after time step $t$ as $\mathbf{h}_{t}^{(m)}$ and $\mathbf{c}_{t}^{(m)}$, respectively. Their initial values $\mathbf{h}_{0}^{(m)}$ and $\mathbf{c}_{0}^{(m)}$  are initialized by  
$\mathbf{h}_{0}^{(m)} = \mathbf{c}_{0}^{(m)} = \mathbf{W}_0^{(m)} \mathbf{a}_0^{(m)}$,
where $\mathbf{W}_0^{(m)} \in \mathbb{R}^{s \times d_m}$, $s$ is the size of LSTM hidden state, and $d_m$ is the size of $\mathbf{a}_0^{(m)}$. At time step $t$, the transition of the $m$-th review component is computed as
\begin{align}
	\left[\mathbf{h}_t^{(m)}, \mathbf{c}_t^{(m)}\right] &= \text{LSTM}_t^{(m)}\left(\mathbf{H}_t,  f_{\textrm{att-fusion-I}}^{(m,t)}\left(A^{(m)}, \mathbf{h}_{t-1}^{(m)}\right)\right),
\end{align}
where 
\begin{align}\label{H_t}
	\mathbf{H}_t &=	\left[ \begin{array}{c}
\mathbf{h}_{t-1}^{(1)}  \\
\vdots \\
\mathbf{h}_{t-1}^{(M)}  \\
\end{array} \right]
\end{align}
is the concatenation of hidden states of all review components at the previous time step, $f_{\textrm{att-fusion-I}}^{(m,t)}(\cdot, \cdot)$ is the attention model for the $m$-th review component, and $\text{LSTM}_t^{(m)}(\cdot, \cdot)$ is the LSTM unit used by the $m$-th review component at time step $t$.  Stage I can be regarded as a grid LSTM \cite{kalchbrenner2015grid} with independent attention mechanisms. In our model, the LSTM unit $\text{LSTM}_t^{(m)}(\cdot, \cdot)$ can be different for different $t$ and $m$. Hence, $M \times T_1$ LSTMs are used in fusion stage I. 
The set of thought vectors generated from the $m$-th component is denoted as:
\begin{align}
	B^{(m)} = \left\{\mathbf{h}_{1}^{(m)}, \mathbf{h}_{2}^{(m)}, \cdots, \mathbf{h}_{T_1}^{(m)}\right\}.
\end{align}

In fusion stage I, the interactions among review components are realized via Eq.~\eqref{H_t}. The vector $\mathbf{H}_t$ contains the hidden states of all the components after time step $t-1$ and is shared as the input among them. Hence, each component is aware of the states of the other components and thus can absorb complementary information from $\mathbf{H}_t$. This hidden state sharing mechanism provides a way for the review component to communicate with each other, which facilitates the generation of thought vectors.

\subsection{Fusion Stage II}
The hidden state and memory cell of fusion stage II are initialized with $\mathbf{h}_{T_1}^{(1)}, \cdots, \mathbf{h}_{T_1}^{(M)}$ and $\mathbf{c}_{T_1}^{(1)}, \cdots, \mathbf{c}_{T_1}^{(M)}$, respectively. We use averaging in our model:
\begin{align}
	\mathbf{h}_{T_1} = \frac{1}{M} \sum_{m=1}^{M} \mathbf{h}_{T_1}^{(m)}, \ \textrm{and} \ \mathbf{c}_{T_1} =\frac{1}{M} \sum_{m=1}^{M} \mathbf{c}_{T_1}^{(m)}.
\end{align}


Fusion stage II combines $M$ sets of thought vectors to form a new one using the multi-attention mechanism. At each time step, the concatenation of context vectors is calculated as:
\begin{align}
	\tilde{\mathbf{z}}_t &=	\left[ \begin{array}{c}
f_{\textrm{att-fusion-II}}^{(1,t)}\left(B^{(1)}, \mathbf{h}_{t-1}\right)  \\
f_{\textrm{att-fusion-II}}^{(2,t)}\left(B^{(2)}, \mathbf{h}_{t-1}\right)  \\
\vdots \\
f_{\textrm{att-fusion-II}}^{(M,t)}\left(B^{(M)}, \mathbf{h}_{t-1}\right)  \\
\end{array} \right],
\end{align}
where $f_{\textrm{att-fusion-II}}^{(m,t)}(\cdot, \cdot)$ is an attention model. Hence, this stage contains $M$ independent attention models. They are all soft attention models, similar to \cite{Xu2016Ask}. With the context vector $\tilde{\mathbf{z}}_t$, the state transition is expressed as:
\begin{align}
	[\mathbf{h}_{t}, \mathbf{c}_{t}] = \text{LSTM}_t(\mathbf{h}_{t-1}, \tilde{\mathbf{z}}_t),
\end{align}
where $\text{LSTM}_{t}$ is the LSTM unit at time step $t$. Please note that all the LSTM units in this stage are also different. 

Fusion stage II can be regarded as review steps~\cite{yang2016review} with $M$ independent attention models, which performs the attention mechanism on the thought vectors yielded in the first stage. It combines and compresses the outputs from stage I and generates only one set of thought vectors. Hence, the generated thought vectors can provide more information for the decoder. 

The hidden states of  fusion stage II are collected to form the thought vector set:
\begin{align}
	C = \left\{ \mathbf{h}_{T_1+1}, \cdots, \mathbf{h}_{T_1 + T_2} \right\},
\end{align}
which will be used as the input of the attention model in the decoder.

\subsection{Decoder}
The decoder translates the information generated by the fusion procedure into  natural sentences. The initial hidden state and memory cell are inherited from the last step of fusion stage II directly. The decoder step in our model is the same as other encoder-decoder models, which is expressed as:
\begin{align}
	[\mathbf{h}_{t}, \mathbf{c}_{t}] = \text{LSTM}_{dec}(\mathbf{H}_{t},  f_{\textrm{att-dec}}\left(C, \mathbf{h}_{t-1}) \right),
\end{align}
where  $\text{LSTM}_{dec}(\cdot, \cdot)$ is the LSTM unit for all the decoder steps, $f_{\textrm{att-dec}}(\cdot, \cdot)$ is the corresponding attention model, $\mathbf{x}_t$ is the word embedding for the input word at the current time step, and $\mathbf{H}_t=	\left[ \begin{array}{c}
\mathbf{x}_t  \\
\mathbf{h}_{t-1}  \\
\end{array} \right]$.

\subsection{Discriminative  Supervision}
We adopt the discriminative supervision in our model to further boost image captioning performance, which is similar to \cite{fang2015captions,yang2016review}. Given a set of thought vectors $V$, a matrix $\mathbf{V}$ is formed by selecting elements from $V$ as column vectors. A score vector $\mathbf{s}$ of words is then calculated as
\begin{align}
	\mathbf{s} = \text{Row-Max-Pool}\left(\mathbf{W} \mathbf{V}\right),
\end{align}
where $\mathbf{W}$ is a trainable linear transformation matrix and $\text{Row-Max-Pool}(\cdot)$ is a max-pooling operator along the rows of the input matrix. The $i$-th element of $\mathbf{s}$ is denoted as $s_{i}$, which represents the score for the $i$-th word. Adopting a multi-label margin loss, we obtain the loss function fulfilling discriminative supervision  as:
\begin{align}
	\mathcal{L}_{d}\left( V \right) = \sum_{j \in \mathcal{W}} \sum_{i \notin \mathcal{W}} \max\left(0, 1-\left(s_{j} - s_{i}\right)\right),
\label{eq_disc}
\end{align}
where $\mathcal{W}$ is the set of all frequent words in the current caption. In this paper, we only consider the 1,000 most frequent words in the captions of the training set.

By considering both the discriminative supervision loss in Eq.~(\ref{eq_disc}) and the captioning loss in Eq.~(\ref{obj_lm}), the complete loss function of our model is expressed as:
\begin{align}
	\mathcal{L}_{all} = \mathcal{L} + \frac{\lambda}{M+1} \left(\mathcal{L}_{d}(C) + \sum_{m=1}^{M} \mathcal{L}_{d}\left(B^{(m)}\right) \right),
\end{align}
where $\lambda$ is a trade-off parameter, and $B^{(m)}$ and $C$ are sets of thought vectors from fusion stages I and II, respectively.

\section{Experiments}

\subsection{Dataset}
The MSCOCO dataset\footnote{http://mscoco.org/} \cite{lin2014microsoft} is the largest benchmark dataset for the image captioning task, which  contains 82,783, 40,504, and 40,775 images for training, validation, and test, respectively. This dataset is challenging, because most images contain multiple objects in the context of complex scenes. Each image in this dataset is associated with five captions annotated by human. For offline evaluation, we follow the conventional evaluation procedure \cite{jmun_aaai2017,yao2016boosting,yang2016review}, and employ the same data split as in \cite{karpathy2015deep},  which contains  5,000 images for validation, 5,000 images for test, and 113,287 images for training.  For online evaluation on the MSCOCO evaluation server, we add the testing set into the training set to form a larger training set.

For the captions, we discard all the non-alphabetic characters, transform all letters into lowercase, and tokenize the captions using white space. Moreover, all the words with the occurrences less than 5 times are replaced by the unknown token {\tt<UNK>}. Thus a vocabulary consisting of 9,487 words is finally constructed. 


\subsection{Configurations and Settings}

For the experiments, we use ResNet~\cite{he2016deep}, DenseNet~\cite{Huang_2017_CVPR}, Inception-V3~\cite{szegedy2016rethinking}, Inception-V4, and Inception-ResNet-V2 \cite{szegedy2017inception} as encoders to extract 5 groups of representations. Each group of representations contains a global feature vector and a set of subregion feature vectors. The  outputs of the last convolution layer (before pooling) are extracted as subregion features. For Inception-V3, the output of the last fully connected layer is used as the global feature vector. For the other CNNs, the means of subregion representations are regarded as the global feature vectors. The parameters for encoders are fixed during the training procedure. Since reinforcement learning (RL) has become a common method to boost image captioning performance~\cite{ranzato2015sequence,rennie2016self,Ren_2017_CVPR,Liu_2017_ICCV,anderson2017bottom}, we first train our model with cross-entropy loss and fine-tune the trained model with CIDEr optimization using reinforcement learning~\cite{rennie2016self}. The performance of models trained with both cross-entropy loss and CIDEr optimization is reported and compared.

When training with cross-entropy loss, the scheduled sampling~\cite{bengio2015scheduled}, label-smoothing regularization (LSR)~\cite{szegedy2016rethinking}, dropout, and early stopping are adopted. For scheduled sampling, the probability of sampling a token from model is $\min(0.25, \frac{epoch}{100})$, where $epoch$ is the number of passes sweeping over training data. For LSR, the prior distribution over labels is uniform distribution and the  smoothing parameter is set to 0.1. Dropout is only applied on the hidden states and the probability is set to 0.3 for all LSTM units. We terminate the training procedure, if the evaluation measurement on validation set, specifically the CIDEr, reaches the maximum value.  When training with RL \cite{rennie2016self}, only dropout and early stopping are used.

The hidden state size is set as 512 for all LSTM units in our model. The parameters of LSTM are initialized with uniform distribution in $[-0.1, 0.1]$. The Adam \cite{kingma2014adam} is applied to optimize the network with the learning rate setting as $5\times 10^{-4}$ and decaying every 3 epochs by a factor 0.8 when training with cross-entropy loss. 
Each mini-batch contains 10 images. 
For RL training, the learning rate is fixed as  $5\times 10^{-5}$. For training with cross-entropy loss, the weight of discriminative supervision $\lambda$ is set to 10. And discriminative supervision is not used for training with RL. To improve the performance,  data augmentation is adopted. Both flipping and cropping strategies are used. We crop $90\%$ of width and height at the four corners. Hence,  $10\times$ images are used for training.

For sentence generation in testing stage, there are two common strategies. The first one is greedy search, which  chooses the word with maximum probability at each time step and sets it as LSTM input for next time step until the end-of-sentence sign is emitted or the maximum length of sentence is reached. The second one is the beam search strategy which selects the top-$k$ best sentences at each time step and considers them as the candidates to generate new top-$k$ best sentences at the next time step. Usually beam search provides better performance for models trained with cross-entropy loss. For model trained with RL, beam search  and greedy search generate similar results. But greedy search is faster than beam search. Hence, for models trained with RL, we use greedy search to generate captions.

\subsection{Performance and Analysis}

We compare our proposed RFNet with the state-of-the-art approaches on image captioning, including Neural Image Caption (NIC) \cite{vinyals2015show},  Attribute LSTM \cite{wu2016value},  LSTM-A3 \cite{yao2016boosting}, Recurrent Image Captioner (RIC) \cite{hliu_ric2016}, Recurrent Highway Network (RHN)~\cite{jgu_rhn2016},  Soft Attention model \cite{xu2015show}, Attribute Attention model~\cite{you2016image}, Sentence Attention model \cite{zhou2016image}, ReviewNet \cite{yang2016review}, Text Attention model \cite{jmun_aaai2017}, Att2in model~\cite{rennie2016self}, Adaptive model~\cite{lu2016knowing},  and Up-Down model~\cite{anderson2017bottom}. Please note that the encoder of Up-Down model is not a CNN pre-trained on ImageNet dataset~\cite{deng2009imagenet}.  It used Faster R-CNN~\cite{ren2015faster} trained on Visual Genome~\cite{krishnavisualgenome} to encode the input image. 
\paragraph{Evaluation metrics.}  
Following the standard evaluation process, five types of metrics are used for performance comparisons, specifically the BLEU \cite{papineni2002bleu},  METEOR \cite{banerjee2005meteor}, ROUGE-L~\cite{lin2004rouge},  CIDEr \cite{vedantam2015cider}, and SPICE~\cite{anderson2016spice}. These metrics measure the similarity between generated sentences and the ground truth sentences from specific viewpoints, \textit{e.g,} n-gram occurrences, semantic contents.  We use the official MSCOCO caption evaluation scripts\footnote{https://github.com/tylin/coco-caption} and the source code of SPICE\footnote{https://github.com/peteanderson80/coco-caption} for the performance evaluation. 

\paragraph{Performance comparisons.}

\begin{table}[t]
\centering
\scriptsize
\caption{Performance comparisons on the test set of Karpathy's split~\cite{karpathy2015deep}. All image captioning models are trained with the cross-entropy loss.  The results are obtained using beam search with beam size 3. $^{\Sigma}$ indicates an ensemble, $^{\dagger}$ indicates a different data split, and $(-)$ indicates that the  metric is not provided. All values are reported in percentage (\%), with the highest value of each entry highlighted in boldface. }
\label{xe_result}
\begin{tabular}{l|c|c|c|c|c|c|c|c}
\toprule
Model & BLEU-1 & BLEU-2 & BLEU-3 & BLEU-4& METEOR & ROUGE-L & CIDEr&SPICE\\
  \midrule
Soft Attention~\cite{xu2015show} &70.7 &49.2 &34.4&24.3&23.9  &-&-& -\\
ReviewNet~\cite{yang2016review} &- &-&-&29.0 &23.7&- &88.6&- \\
LSTM-A3~\cite{yao2016boosting}&73.5 &56.6 &42.9 &32.4 &25.5 &53.9 &99.8 &18.5\\
Text Attention \cite{jmun_aaai2017}  &74.9 &58.1 &43.7&32.6&25.7&-&102.4&-\\
Attribute LSTM \cite{wu2016value} &74.0 &56.0&42.0&31.0&26.0&-&94.0&-\\
RIC \cite{hliu_ric2016}  & 73.4 & 53.5& 38.5&29.9&25.4&-&-&-\\
RHN \cite{jgu_rhn2016} &72.3 &55.3 &41.3 &30.6&25.2&-&98.9&-\\
Adaptive \cite{lu2016knowing}$^{\dagger}$ & 74.2 & 58.0 & 43.9&33.2 &26.6  &- &108.5& - \\
Att2in \cite{rennie2016self} &-&-&-&31.3&26.0&54.3&101.3 &-\\
Up-Down~\cite{anderson2017bottom} &77.2 &- &-&36.2 &27.0  &56.4 &113.5& 20.3 \\
  \midrule
 Sentence Attention \cite{zhou2016image}$^{\Sigma}$ &71.6 & 54.5&40.5&30.1&24.7  &-&97.0&-\\
 Attribute Attention \cite{you2016image}$^{\Sigma}$   &70.9 &53.7 &40.2&30.4 & 24.3  &-&-&-\\
 NIC \cite{vinyals2015show}$^{\dagger}$$^{\Sigma}$   & - & - & -&32.1&25.7  &-&99.8&-\\
 Att2in~\cite{rennie2016self}$^{\Sigma}$  & - & - & -& 32.8&26.7  &55.1&106.5&- \\
ReviewNet~\cite{yang2016review}$^{\Sigma}$   &76.7 &60.9 &47.3 &36.6 &27.4 &56.8 &113.4 &20.3\\
  \midrule
RFNet  &76.4 &60.4 &46.6  &35.8 &27.4 &56.5 &112.5 &20.5\\ 
RFNet$^{\Sigma}$  &\textbf{77.4} &\textbf{61.6}&\textbf{47.9} &\textbf{37.0} &\textbf{27.9} &\textbf{57.3} &\textbf{116.3} &\textbf{20.8}\\
\bottomrule
\end{tabular}
\vspace{-0.3cm}
\end{table}

 \begin{table}[t]
\centering
\scriptsize
\caption{Performance comparisons on the test set of Karpathy's split~\cite{karpathy2015deep}. All models are fine-tuned with RL. $^{\Sigma}$ indicates an ensemble, $^{\dagger}$ indicates a different data split, and  $(-)$ indicates that the  metric is not provided. All values are reported in percentage (\%), with the highest value of each entry highlighted in boldface. The results are obtained using greedy search.}
\label{rl_result}
\begin{tabular}{l|c|c|c|c|c|c|c|c}
\toprule
Model & BLEU-1 & BLEU-2 & BLEU-3 & BLEU-4& METEOR & ROUGE-L & CIDEr & SPICE\\
\midrule
Att2all~\cite{rennie2016self}  & - & - & -&34.2 &26.7  &55.7&114.0&- \\
Up-Down \cite{anderson2017bottom}  & 79.8 & - & -& 36.3 &27.7  &56.9&120.1&21.4 \\
Att2all~\cite{rennie2016self}$^{\Sigma}$  & - & - & -&35.4 &27.1  &56.6&117.5&- \\
ReviewNet~\cite{yang2016review}$^{\Sigma}$ &79.6 &63.6 &48.7 &36.4 &27.7 &57.5 &121.6&21.0 \\ 
  \midrule
RFNet&79.1 &63.1 &48.4 &36.5  &27.7 &57.3  &121.9  & 21.2\\
RFNet$^{\Sigma}$ &\textbf{80.4} &\textbf{64.7} &\textbf{50.0} &\textbf{37.9} &\textbf{28.3} &\textbf{58.3} &\textbf{125.7} &\textbf{21.7}  \\ 
\bottomrule
\end{tabular}
\end{table}

\begin{table}[!t]
\centering
\tiny
 \caption{Performance of different models on the MSCOCO evaluation server. All values are reported in percentage (\%), with the highest value of each entry  highlighted in boldface.}
 \label{online_result}
 \begin{tabular}{l| c|c| c|c| c|c| c|c| c|c| c|c| c|c| c|c}
\toprule
&\multicolumn{2}{c|}{BLEU-1}
&\multicolumn{2}{c|}{BLEU-2} 
&\multicolumn{2}{c|}{BLEU-3} 
&\multicolumn{2}{c|}{BLEU-4} 
&\multicolumn{2}{c|}{METEOR} 
&\multicolumn{2}{c|}{ROUGE-L} 
&\multicolumn{2}{c|}{CIDEr} 
&\multicolumn{2}{c}{SPICE} 
\\
\cmidrule(r){2-3}
\cmidrule(l){4-5}
\cmidrule(l){6-7} 
\cmidrule(l){8-9} 
\cmidrule(l){10-11} 
\cmidrule(l){12-13} 
\cmidrule(l){14-15} 
\cmidrule(l){16-17}
Method &C5& C40    &C5&C40  &C5&C40 &C5&C40 &C5&C40 &C5&C40 &C5&C40 &C5&C40     \\
\midrule
NIC~\cite{vinyals2016show} &71.3 &89.5 &54.2 &80.2 &40.7 &69.4 &30.9 &58.7 &25.4 &34.6 &53.0 &68.2 &94.3 &94.6 &18.2 &63.6 \\
Captivator~\cite{fang2015captions} &71.5 &90.7 &54.3 &81.9 &40.7 &71.0 &30.8 &60.1 &24.8 &33.9 &52.6 &68.0 &93.1 &93.7 &18.0 &60.9 \\
M-RNN~\cite{mao2014deep}  &71.6 &89.0 &54.5 &79.8 &40.4 &68.7 &29.9 &57.5 &24.2 &32.5 &52.1 &66.6 &91.7 &93.5 &17.4 &60.0 \\
LRCN~\cite{donahue2015long}  &71.8 &89.5 &54.8 &80.4 &40.9 &69.5 &30.6 &58.5 &24.7 &33.5 &52.8 &67.8 &92.1 &93.4 &17.7 &59.9 \\
Hard-Attention \cite{xu2015show}  &70.5 &88.1 &52.8 &77.9 &38.3 &65.8 &27.7 &53.7 &24.1 &32.2 &51.6 &65.4 &86.5 &89.3 &17.2 &59.8 \\
ATT-FCN~\cite{you2016image}  &73.1 &90.0 &56.5 &81.5 &42.4 &70.9 &31.6 &59.9 &25.0 &33.5 &53.5 &68.2 &94.3 &95.8 &18.2 &63.1 \\
ReviewNet~\cite{yang2016review}  &72.0 &90.0 &55.0 &81.2 &41.4 &70.5 &31.3 &59.7 &25.6 &34.7 &53.3 &68.6 &96.5 &96.9 &18.5 &64.9 \\
LSTM-A3~\cite{yao2016boosting}  &78.7 &93.7 &62.7 &86.7 &47.6 &76.5 &35.6 &65.2 &27.0 &35.4 &56.4 &70.5 &116.0 &118.0 &-&-\\
Adaptive~\cite{lu2016knowing} &74.8 &92.0 &58.4 &84.5 &44.4 &74.4 &33.6 &63.7 &26.4 &35.9 &55.0 &70.5 &104.2 &105.9 &19.7 &67.3 \\
PG-BCMR~\cite{Liu_2017_ICCV} &75.4 &91.8 &59.1 &84.1 &44.5 &73.8 &33.2 &62.4 &25.7 	&34.0 &55.0 &69.5 &101.3 &103.2 &- &-  \\
Att2all~\cite{rennie2016self} &78.1 &93.7 &61.9 &86.0 &47.0 &75.9 &35.2 &64.5 &27.0 &35.5 &56.3 &70.7 &114.7 &116.7 & 20.7 & 68.9 \\
Up-Down~\cite{anderson2017bottom} &80.2&\textbf{95.2} &64.1 &88.8 &49.1 &79.4 &36.9 &68.5 &27.6 &36.7 &57.1 &72.4 &117.9 &120.5 &21.5	 & 71.5\\
\hline
RFNet &\textbf{80.4} &95.0 &\textbf{64.9} &\textbf{89.3} &\textbf{50.1} &	\textbf{80.1} &\textbf{38.0} &\textbf{69.2} &\textbf{28.2} &\textbf{37.2} &\textbf{58.2} &\textbf{73.1} &\textbf{122.9} &\textbf{125.1} & -&-  \\ 
\bottomrule
\end{tabular}
\vspace{-0.2cm}
\end{table}

The performance of models trained with cross-entropy loss is shown in Table~\ref{xe_result}, including the performance of single model and ensemble of models. First, it can be observed that our single model RFNet  significantly outperformed the existing image captioning models, except the Up-Down model~\cite{anderson2017bottom}. Our RFNet performed inferiorly to Up-Down in \mbox{BLEU-1}, \mbox{BLEU-4}, and CIDEr, while superiorly to Up-down in METEOR, \mbox{ROUGGE-L}, and SPICE. However, the encoder of \mbox{Up-Down} model was pre-trained on ImageNet dataset and fine-tuned on Visual Genome~\cite{krishnavisualgenome} dataset. The Visual Genome dataset is heavily annotated with objects, attributes and region descriptions and 51K images are extracted from MSCOCO dataset. Hence, the encoder of \mbox{Up-Down} model is trained with far more information than the CNNs trained on ImageNet. With the recurrent fusion strategy, RFNet can extract useful information from different encoders to remedy the lacking of information about objects and attributes in the representations.


Moreover,  we can observe that our \textit{single RFNet model} performed significantly better than other ensemble models, such as NIC, Att2in, and behaved comparably with ReviewNet$^{\Sigma}$ which is an ensemble of 40 ReviewNets (8 models for each CNNs). But RFNet$^{\Sigma}$, an ensemble of 4 RFNets, significantly outperformed all the ensemble models.

The performance comparisons with RL training are presented in Table~\ref{rl_result}. We compared RFNet with Att2all \cite{rennie2016self}, Up-Down model \cite{anderson2017bottom}, and ensemble of ReviewNets. For our method, the performance of single model and ensemble of 4 models are provided. We can see that our RFNet outperformed other methods. For online evaluation, we used the ensemble of 7 models and the comparisons are provided in Table~\ref{online_result}. We can see that our RFNet still achieved the best performance. The C5 and C40 CIDEr scores were improved by 5.0 and 4.6, compared to the state-of-the-art Up-Down model~\cite{anderson2017bottom}. 



\paragraph{Ablation study of fusion stages I and II.}
To study the effects of the two fusion stages, we present the performance of the following models:

\begin{itemize}
	\item {$\textrm{RFNet}_{-\textrm{I}}$} denotes RFNet without fusion stage I, with only the fusion stage II preserved.  The global representations are concatenated to form one global representation and multi-attention mechanisms are performed on the subregion representation from the multiple encoders. The rest is the same with RFNet. 
	\item {$\textrm{RFNet}_{-\textrm{II}}$} denotes RFNet without fusion stage II. Multiple attention models are employed in the decoder and the rest is the same as RFNet. 
	\item {$\textrm{RFNet}_{-\textrm{inter}}$} denotes RFNet without the interactions in fusion stage I. Each component in the fusion stage I is independent. Specifically, at each time step, the input of the $m$-th component is just $\mathbf{h}_{t-1}^{(m)}$, and it is unaware of the hidden states of the other components.
\end{itemize}
The CIDEr scores on the test set of the Karpathy's split are presented in Table~\ref{ablation}. We can see that both the two-stage structure and the interactions in the first stage are important for our model. With the interactions, the quality of thought vectors in the first stage can be improved. With the two-stage structure, the thought vectors in the first stage can be refined and compressed into more compact and informative set of thought vectors  in the second stage. Therefore, with the specifically designed recurrent fusion strategy, our proposed RFNet provides the best performance.

\begin{table}[h]
\centering
\caption{Ablation study of fusion stages I and II in our RFNet. The CIDEr scores are obtained using beam search with beam size 3 on the test set of the Karpathy's split.}
\label{ablation}
\begin{tabular}{p{1.5cm}<{\centering}|p{2cm}<{\centering}|p{2cm}<{\centering}|p{2cm}<{\centering}|p{2cm}<{\centering}}
\toprule
 &$\textrm{RFNet}_{-\textrm{I}}$ & $\textrm{RFNet}_{-\textrm{II}}$ & $\textrm{RFNet}_{-\textrm{inter}}$ &RFNet  \\
    \midrule
CIDEr &108.9 &111.5 &112.1 &112.5 \\
\bottomrule
\end{tabular}
\end{table}
\vspace{-0.5cm}


\paragraph{Effects of discriminative supervision.}
We examined the effects of the discriminative supervision with different values of $\lambda$. First, it can be observed that introducing discriminative supervision properly can help improve the captioning performance, with 107.2 ($\lambda$=1) and 107.3 ($\lambda$=10)  vs. 105.2 ($\lambda$=0). However, if $\lambda$ is too large, \textit{e.g.}, 100, the discriminative supervision will degrade the corresponding performance. Therefore, in this paper,  $\lambda$ is set as 10, which provided the best performance. 

\begin{table}[h]
\centering
\caption{CIDEr scores with different $\lambda$ values on the test set. The captions are generated by greedy search.}
\begin{tabular}{p{1.5cm}<{\centering}|p{1.5cm}<{\centering}|p{1.5cm}<{\centering}|p{1.5cm}<{\centering}|p{1.5cm}<{\centering}}
\toprule
$\lambda$ &0 &1 &  10 & 100  \\
  \midrule
CIDEr &105.2 &107.2 &107.3 &104.7 \\
\bottomrule
\end{tabular}
\end{table}
\vspace{-0.5cm}



\section{Conclusions}

In this paper, we proposed a novel Recurrent Fusion Network (RFNet), to exploit complementary information of multiple image representations for image captioning. In the RFNet, a recurrent fusion procedure is inserted between the encoders and the decoder.This recurrent fusion procedure consists of two stages, and each stage can be regarded as a special RNN. In the first stage, each image representation is compressed into a set of thought vectors by absorbing complementary information from the other representations. The generated sets of thought vectors are then compressed into another set of thought vectors in the second stage, which will be used as the input to the attention module of the decoder. The proposed RFNet achieved leading performance on the MSCOCO evaluation server, corroborating the effectiveness of our proposed network architecture.
\clearpage

\bibliographystyle{splncs}
\bibliography{egbib}
\clearpage

\appendix
\section{Appendices}

\subsection{Qualitative Analysis} 
\begin{figure}[h]
 \centering
  \begin{tabular}{ | c | m{5.3cm} | m{5.3cm} | }
\toprule
    Image & Captions generated & Ground truth \\
    \hline 
    \begin{minipage}{.15\textwidth}
      \includegraphics[width=\linewidth, height=0.8\linewidth]{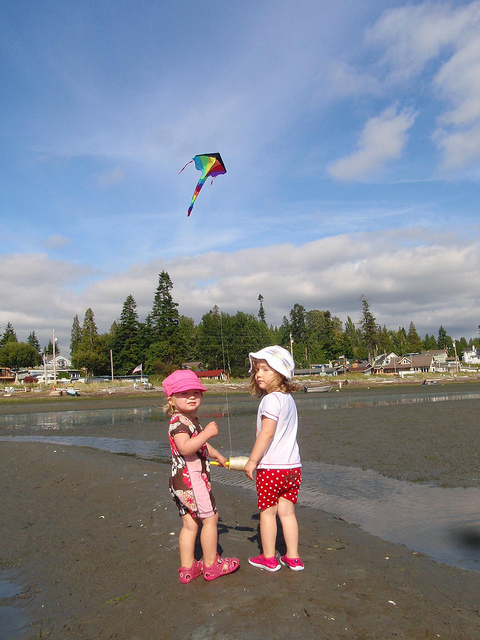}
    \end{minipage} 
    & 
	\begin{minipage}[t]{5.25cm}\scriptsize
        \textbf{Our}: Two children are flying a kite on the beach. \\
        \textbf{ReviewNet-ResNet}: A little girl standing next to a little girl.\\
        \textbf{ReviewNet-DenseNet}: A man and a little girl flying a kite. \\
        \textbf{ReviewNet-Inception-V3}: A couple of kids standing on top of a beach.
    \end{minipage}
    &
    \begin{minipage}{5.25cm} \scriptsize 
    \vspace{0.2cm}
1. Two children flying a kite near the beach. \\
2. Two young children flying a kite on the beach. \\
3. Two young children fly their kite in the blue sky. \\
4. Two children holding a kite flying in the air. \\
5. A couple of kids flying a kite on top of a beach. \\ 
    \end{minipage} \\

    \hline 
    \begin{minipage}{.15\textwidth}
      \includegraphics[width=\linewidth, height=0.8\linewidth]{433243.jpg}
    \end{minipage}
    & 
	\begin{minipage}[t]{5.25cm}\scriptsize
        \textbf{Our}: A zebra and a giraffe standing next to each other. \\
        \textbf{ReviewNet-ResNet}: A couple of giraffe standing next to each other. \\
        \textbf{ReviewNet-DenseNet}: A zebra standing in the dirt near a fence. \\
        \textbf{ReviewNet-Inception-V3}: A couple of zebra standing next to each other.
    \end{minipage}
    &
    \begin{minipage}{5.25cm}\scriptsize     \vspace{0.2cm}
1. A zebra and a giraffe eat some hay together. \\
2. A zebra and a giraffe stand next to each other. \\
3. A giraffe and a zebra stand side by side as they eat. \\
4. Giraffe and a zebra standing side by side.  \\
5. A giraffe eats next to a zebra among some rocks. \\
    \end{minipage} \\


        \hline 
    \begin{minipage}{.15\textwidth}
      \includegraphics[width=\linewidth, height=0.8\linewidth]{230008.jpg}
    \end{minipage}
    &
	\begin{minipage}[t]{5.25cm}\scriptsize
        \textbf{Our}: A police officer is sitting on a motorcycle. \\
        \textbf{ReviewNet-ResNet}: A couple of men standing next to a motorcycle. \\
        \textbf{ReviewNet-DenseNet}: A motorcycle is parked on the side of the road. \\
        \textbf{ReviewNet-Inception-V3}: A man on a motorcycle in a parking lot.
            \end{minipage}
    &
    \begin{minipage}{5.25cm}\scriptsize     \vspace{0.2cm}
1. A police officer on a motorcycle sitting at the corner. \\
2. A traffic motorcycle cop waits to give a ticket.\\
3. A police officer is outside on his bike. \\
4. A police man on a motorcycle is idle in front of a bush. \\
5. Police officer laying a speed trap for the purpose of writing a ticket.\\
    \end{minipage} \\
    

    \hline 
    \begin{minipage}{.15\textwidth}
      \includegraphics[width=\linewidth, height=0.8\linewidth]{113836.jpg}
    \end{minipage}
    &
	\begin{minipage}[t]{5.25cm}\scriptsize
        \textbf{Our}: A hotel room with two beds and a lamp. \\
        \textbf{ReviewNet-ResNet}: A bedroom with a neatly made bed and two lamps. \\
        \textbf{ReviewNet-DenseNet}: A bedroom with a bed and a lamp. \\
        \textbf{ReviewNet-Inception-V3}: A bedroom with a bed and two lamps.
    \end{minipage}
    &
    \begin{minipage}{5.25cm}\scriptsize     \vspace{0.2cm}
1. A couple of beds sitting next to each other in a room. \\
2. There is a room with two beds in it.  \\
3. There are two small twin beds in this room. \\
4. Two made beds, a nightstand with a lamp, and framed art on the wall. \\
5. A bedroom with two beds and a nightstand with a lamp in between the beds. \\ 
    \end{minipage} \\

\bottomrule
  \end{tabular}
  \caption{Examples of captions generated by our method and Review Net models taking different CNNs as encoders, as well as the corresponding ground truths.}
  \label{caption-example}
\end{figure}

\begin{figure}[h]
 \centering
  \begin{tabular}{ | c | m{5.3cm} | m{5.3cm} | }
\toprule 
    Image & Captions generated & Ground truth \\
    \hline 
    
   \hline 
    \begin{minipage}{.15\textwidth}
      \includegraphics[width=\linewidth, height=0.8\linewidth]{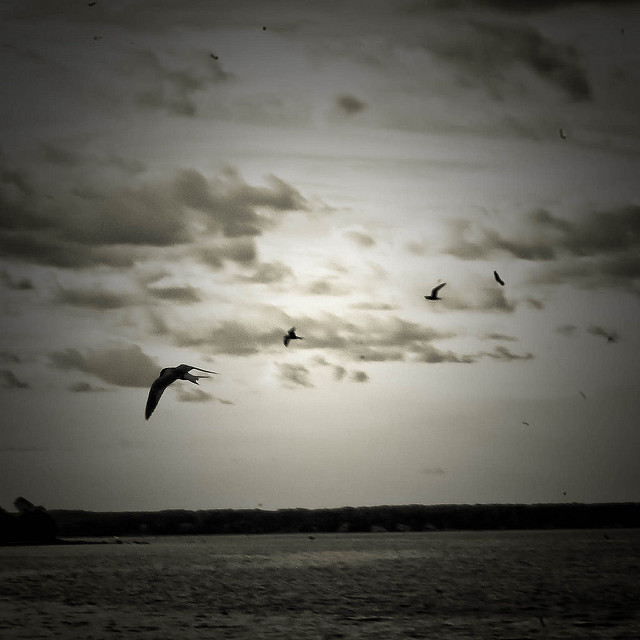}
    \end{minipage}
    &
	\begin{minipage}[t]{5.25cm}\scriptsize
        \textbf{Our}: A flock of birds flying over a body of water.\\
        \textbf{ReviewNet-ResNet}: A bird flying over a body of water. \\
       	\textbf{ReviewNet-DenseNet}: A bird flying over a body of water. \\
       	 \textbf{ReviewNet-Inception-V3}: A bird flying over a body of water.
    \end{minipage}
    &
    \begin{minipage}{5.25cm}\scriptsize
1. A flock of small birds flying in the sky over the water. \\
2. Several birds that are flying together over a body of water. \\
3. A flock of birds flying over a field. \\
4. A black and white image showing birds flying over a body of water. \\
5. A group of birds flying over the water. 
    \end{minipage}\\
    
       \hline 
    \begin{minipage}{.15\textwidth}
      \includegraphics[width=\linewidth, height=0.8\linewidth]{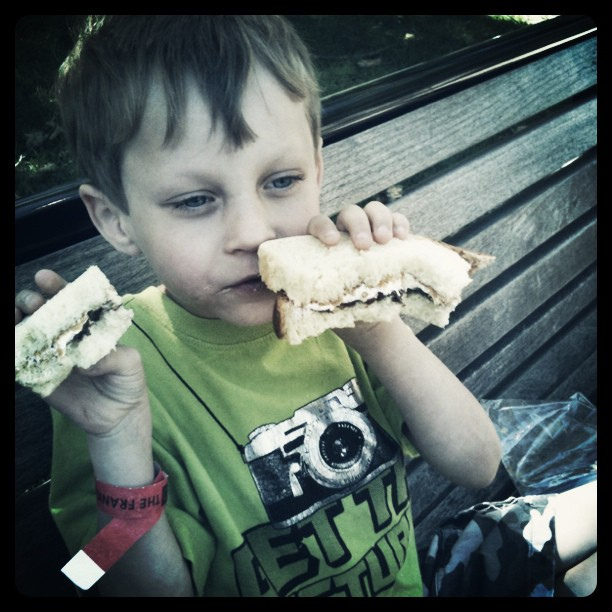}
    \end{minipage}
    &
	\begin{minipage}[t]{5.25cm}\scriptsize
        \textbf{Our}: A little boy sitting on a bench eating a sandwich. \\
        \textbf{ReviewNet-ResNet}: A young boy eating a doughnut in front of a building. \\ 
        \textbf{ReviewNet-DenseNet}: A young boy is eating a hot dog. \\
        \textbf{ReviewNet-Inception-V3}: A young boy eating a piece of cake.
    \end{minipage}
    &
    \begin{minipage}{5.25cm}\scriptsize 
1. A distant airplane flying between two large buildings. \\
2. A couple of tall buildings with a large jetliner between them. \\
3. Commercial airliner flying between two tall structures on clear day. \\
4. Two tall buildings and a plane in between them. \\
5. An airplane is seen between two identical skyscrapers. 
    \end{minipage} \\

\hline 
    \begin{minipage}{.15\textwidth}
      \includegraphics[width=\linewidth, height=0.8\linewidth]{94157.jpg}
    \end{minipage}
    &
	\begin{minipage}[t]{5.25cm}\scriptsize
        \textbf{Our}: A large clock tower in front of a building. \\
        \textbf{ReviewNet-ResNet}: A large building with a clock on it. \\
        \textbf{ReviewNet-DenseNet}: A tall clock tower towering over a city. \\
        \textbf{ReviewNet-Inception-V3}: A very tall clock tower with a sky background. \\
    \end{minipage}
    &
    \begin{minipage}{5.25cm}\scriptsize
1. A group of people in a park with a tall clock tower in the background. \\
2. A clock tower in front of a building. \\
3. Large clock tower sitting in front of a park. \\ 
4. A large, ornate clock tower rises above another skyscraper. \\
5. A large brick clock tower stands high above the trees. 
    \end{minipage} \\

%
    
    \hline 
    \begin{minipage}{.15\textwidth}
      \includegraphics[width=\linewidth, height=0.8\linewidth]{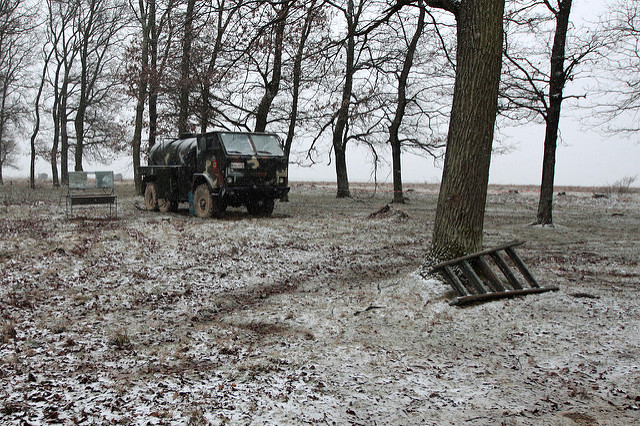}
    \end{minipage}
    &
	\begin{minipage}[t]{5.25cm}\scriptsize
        \textbf{Our}: A truck is parked in the middle of a forest. \\
        \textbf{ReviewNet-ResNet}: A truck parked on the side of a dirt road. \\
        \textbf{ReviewNet-DenseNet}: A truck is parked on the side of the road. \\
        \textbf{ReviewNet-Inception-V3}: A truck driving down a dirt road next to a forest.
    \end{minipage}
    &
    \begin{minipage}{5.25cm}\scriptsize
1. A truck is parked in the middle of a forest. \\
2. A large vehicle is parked in a field with trees in it. \\
3. A large vehicle in a deserted park covered in snow. \\
4. A truck is parked at a campground with snow on it. \\
5. A truck with the hood open sits in an area that has been lightly dusted with snow and is a wooded area with sparse trees. 
    \end{minipage} \\

    
        \hline 
    \begin{minipage}{.15\textwidth}
      \includegraphics[width=\linewidth, height=0.8\linewidth]{244215.jpg}
    \end{minipage}
    &
	\begin{minipage}[t]{5.25cm}\scriptsize
        \textbf{Our}: Two dogs sitting in front of a wooden structure. \\
        \textbf{ReviewNet-ResNet}: A dog standing next to a pile of garbage. \\
        \textbf{ReviewNet-DenseNet}: A brown dog standing next to a yellow fire hydrant. \\
        \textbf{ReviewNet-Inception-V3}: A couple of dogs standing next to each other.
    \end{minipage}
    &
    \begin{minipage}{5.25cm}\scriptsize
1. A couple of dog statues next to some pallets and fire extinguishers. \\
2. A couple of dogs sitting in front of a metal structure. \\
3. There are two fake dogs outside a warehouse. \\
4. Statues of two dogs sitting in front of a building. \\ 
5. Two fake dogs sit in front of a building. 
    \end{minipage} \\

    
    \hline 
    \begin{minipage}{.15\textwidth}
      \includegraphics[width=\linewidth, height=0.8\linewidth]{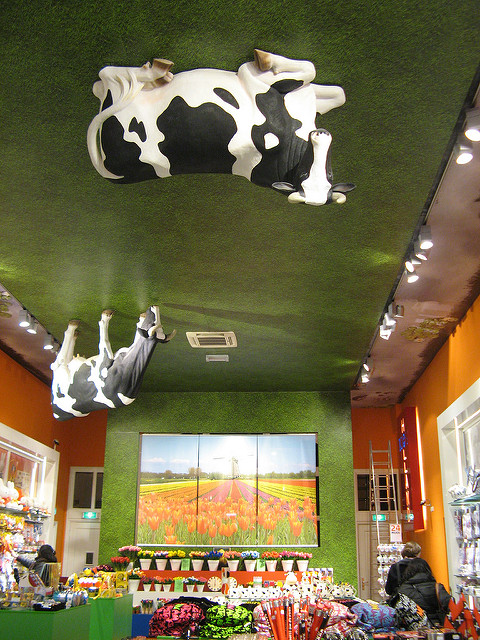}
    \end{minipage}
    &
	\begin{minipage}[t]{5.25cm}\scriptsize
        \textbf{Our}: A picture of a cow that is in the air. \\
        \textbf{ReviewNet-ResNet}: A statue of a cow in a room. \\
        \textbf{ReviewNet-DenseNet}: A display of a store with lots of different types of items. \\
        \textbf{ReviewNet-Inception-V3}: A room filled with lots of different colored kites.
    \end{minipage}
    &
    \begin{minipage}{5.25cm}\scriptsize
1. A room with cows appearing to be on the grass on the ceiling. \\
2. A very colorful room with models of cows afixed to the ceiling. \\
3. A large cow that is attached to the cieling. \\
4. Cows on display on top of codling with people far down below. \\
5. Two statues of cows set on the roof. 
    \end{minipage} \\
    
    \hline 
    \begin{minipage}{.15\textwidth}
      \includegraphics[width=\linewidth, height=0.8\linewidth]{465973.jpg} 
    \end{minipage}
    &
	\begin{minipage}[t]{5.25cm}\scriptsize
        \textbf{Our}: A man riding on the back of a cow. \\
        \textbf{ReviewNet-ResNet}: A man riding a horse on a dirt road. \\
        \textbf{ReviewNet-DenseNet}: A man riding on the back of a horse. \\
        \textbf{ReviewNet-Inception-V3}: A man is riding a horse in the dirt.
    \end{minipage} 
    &
    \begin{minipage}{5.25cm}\scriptsize
1. A man in a hat riding a horse. \\
2. A man riding a horse as some people watch. \\
3. A horse being ridden with a man in a large hat. \\
4. A man in a sombrero rides a black horse in a dirt ring. \\
5. A man is riding a horse in a dirt enclosure. 
    \end{minipage} \\

\bottomrule
  \end{tabular}
  \caption{More examples.}
  \label{caption-example_2}
\end{figure}

To gain a qualitative understanding of our model, some examples of captions generated by our model and ReviewNet are shown in Fig.~\ref{caption-example}. We only provide the results of ReviewNets with ResNet, DenseNet and Inception-V3 as encoders. We can observe that the improved performance of our model is mostly attributed to the improvement in the recognition of salient objects. For example, our model recognized ``\texttt{two children}'', ``\texttt{a kite}'' and ``\texttt{beach}'' in the first image. But ReviewNet-ResNet failed to recognize ``\texttt{beach}'', ReviewNet-DenseNet wrongly recognized ``\texttt{a man}'' and ReviewNet-Inception-V3 failed to recognize ``\texttt{kite}''. Similar results can also be observed in the other cases. Like the ``\texttt{zebra}'' and ``\texttt{giraffe}'' in the second image, the ``\texttt{police officer}'' in the third image and the ``\texttt{two beds}'' and ``\texttt{a lamp}'' in the last image. The Review Net with only one kind of CNN does not recognize the salient objects as our model does. Hence, by the recurrent fusion process, our model is good at combining multiple representations and finding the salient objects neglected in individual representations. More examples can be found in Fig~\ref{caption-example_2}.

\subsection{Ablation Study on Different CNNs.} 

We perform the ablation studies to show the effects of different CNNs used in our experiments, with the results shown in Table~\ref{tab}. We can see that generally speaking, ResNet contributes the most while DenseNet contributes the least to the final image captioning performance.

\begin{table}
\centering
\caption{CIDEr scores with different CNNs on the validation set. The captions are generated by greedy search. Features from ResNet, Inception-V4, Inception-V3, DenseNet and Inception-ResNet-V2 are denoted as F1, F2, F3, F4 and F5, respectively.
}\label{tab}
\begin{tabular}{l|c|c|c|c|c|c|c|c|c}
\toprule
CNNs & F1+F2 &F1+F2+F3 &F1+F2+F3+F4 & All & All-F1 &All-F2 &All-F3 & All-F4 & All-F5  \\
  \midrule
CIDEr &106.35 & 106.85 &106.72  &107.15 &104.8 &106.80 &106.90  &107.10 &106.72 \\
\bottomrule
\end{tabular}
\end{table}
\clearpage

\end{document}